# Artificial Intelligence in the Low-Level Realm – A Survey

Vahid Mohammadi Safarzadeh

Hamed Ghasr Loghmani

*Abstract*—Resource-aware machine learning has been a trending topic in recent years, focusing on making ML computational aspects more exploitable by the edge devices in the Internet of Things. This paper attempts to review a conceptually and practically related area concentrated on efforts and challenges for applying ML in the operating systems' main tasks in a low-resource environment. Artificial Intelligence has been integrated into the operating system with applications such as voice or image recognition. However, this integration is only in user space. Here, we seek methods and efforts that exploit AI approaches, specifically machine learning, in the OSes' primary responsibilities. We provide the improvements that ML can bring to OS to make them more trustworthy. In other words, the main question to be answered is how AI has played/can play a role directly in improving the traditional OS kernel main tasks. Also, the challenges and limitations in the way of this combination are provided.

## I. Introduction

Artificial Intelligence and its famous subcategory, Machine Learning, is the current trend in science. It has affected many scientific and technological applications such as medicine, science, and industry. However, it is interesting that OSes' low-level operations as the basics for all computing jobs have remained almost untouched by this revolutionary science. OS algorithms seem very similar to that of twenty years ago with no sign of intelligence in its current meaning; learning and adapting. Although AI does not guarantee an improvement in every application and also the idea of applying it to any traditional procedure and expecting miracles seems naive, the origins of uncertainty and unpredictability that OSes are facing now can be reasonable for AI practitioners to see them as problems to be solved via the available tools of AI.

At first sight, mechanisms such as memory management and process scheduling are the playgrounds that AI can help OS with its current procedures. A computer executes programs. The programs and the data they need to access should be kept, at least partially, in the main memory. A memory management scheme is for managing the presence and replacement of the data in the main memory to maximize the CPU utilization and minimize the response time to users. For the same purposes, the process schedulers choose one process from all ready-to-run or waiting for processes to be executed by the CPU. However, several challenges need to be addressed. OSes are programs that resulted from many years of development. They must remain as light-weight as possible while handling the most critical job in a system in the most deterministic and flawless manner. Its actions also have many parameters, consequences, and players. Therefore, in OSes, as an automated and deterministic mechanism, AI can have a role in ways that do not compromise the system's correctness [1]. Here, our focus is on the kernel space applications of AI in OS. But some applications exploit both user and kernel space, mostly due to the resource limitation in the kernel space.

Algorithms used in OS design are either deterministic or heuristic. A deterministic algorithm produces the same output for specific given input and in any number of runs, going through the same states. On the other hand, a heuristic algorithm offers a procedure to achieve an optimal or near-optimal solution to an optimization problem [2] with which the output may change for the same data in different runs. The heuristic ones are mainly more monitored trial and error methods. For example, for resource scheduling, these heuristics are adjusted through monitoring the changes occurring in the system's performance while increasing/decreasing specific resources for a process [3].

The overwhelming wave of ML computational methods is affecting the lower level of computers. Several concepts such as learned data structures (Indexes) [4], learned database systems (for optimizing queries) [5], learned indexes [6]–[8] and learning memory hit pattern [9] are introduced, recently, which shows the applications of AI into the low-level world of computation. The main feature of ML is the ability to exploit experiences of the past for future decisions.

Searching in configuration spaces and predicting the system's future states, timings, or sizes are jobs that OSes are continuously performing. These operations become even more crucial in applications such as storing systems, web servers, High-Performance Computing systems, clouds, and real-time applications.

In this paper, along with analysing some challenges of using AI to enhance OS operations, previous efforts in this area are analysed. Linux's open-source nature makes it a suitable platform to apply AI methods to an OS's different tasks, although the implementation is challenging, as we will encounter in this paper.

There also have been efforts to make hardware architecture more intelligent using ML. For example, in [9], the prefetching operation (in which hardware predicts memory accesses before the request by software based on history) is modelled as a sequence prediction problem. They reported a solution to this challenge using Long Short-Term Memory (LSTM) Recurrent Neural Networks. However, we do not encounter such integrations in this paper.

The kernel, services, and Shell constitute an OS. The kernel is the central part, and other parts can be seen as tools to interact with the kernel and send requests to it. The primary responsibilities of OSes are managing software and hardware as well as providing system services.

Some of these services fulfil users' (processes) requests such as I/O requests, file-system manipulation, or inter-process communications. Others are for increasing the system's efficiency. These services are mainly related to sharing resources among processes or threads using resource allocation algorithms; we mean CPU cycles, main memory, file storage, or I/O devices [10] .

Another important job of OS (and hardware) is caching. With the increasing popularity of cloud computing, some approaches towards programming have been changed. One of the most important changes was using microservices to write applications instead of the traditional monolithic manner in which all parts of the program were in one piece. In a microservice-based development style, a single application is delivered as several small services developed independently and loosely coupled. As a result, the program becomes highly maintainable and testable. Each service needs to communicate with the user and other services. Obviously, breaking programs into microservices provides new challenges to OSes in handling the requests of these services. Dealing with the massive amount of microservices demands can not be done optimally with traditional operational methods [3].

Machine Learning is a programming approach to use experiences for making better future decisions. Each learning application is divided into two phases: Training to produce a model and Inference based on that model. The first phase is more resource-demanding and time-consuming than the second. A subcategory of ML, Deep Learning, has gained special attention in recent years with the advent of new technologies and the availability of more data. Deep learning is based on the usage of artificial neural networks with representation learning. The performance of machine learning methods is heavily dependent on the choice of data representation (or features) on which they are applied [11]. Representation learning offers methods to automatically find the desirable data representation (or features).

In this paper, we first make a connection between the main goal of AI (ML) methods which is dealing with uncertainties and the operating system's tasks in which the OS faces unpredictable situations. Then, we analyse several reported efforts in the literature in integrating ML with OS. Finally, we encounter some challenges and possible solutions for this integration. The conclusion and future of the idea of embedding ML into OS come in the last section. In addition, since the subject of this paper is somehow an interdisciplinary topic in computer science, we provide enough information and definitions to make it self-contained so that any expert from both domains (AI or OS) can use it smoothly.

## II. Origins of Uncertainty in Operating System

OS designers try to make their systems as predictable as possible. By predictability, we mean the system's ability to evaluate each task's timing and resource properties. One smart idea is for OS to provide the exact amount of resources that every algorithm requires when it is called to perform. This behaviour will optimize the OS (server) load and reduce the energy that it needs. On the other hand, such predictability may have security drawbacks because it simplifies how malware designers can communicate with OS [12], [13]. Therefore, any approach (including ML approaches) should consider such a threat to the final system.

Some sources of uncertainty in OSes like the uncertainty in reading timestamps resulted from Linux Kernel overhead, on which the load of the system and CPU clock speed has a direct impact [14]. Although such uncertainties may not be harmful in desktop or even traditional server usages of OSes, despite the energy deficiency that they cause, they are crucial in real-time and embedded OSes. The majority of embedded devices tend to fulfil a specific task using real-time operating systems. A real-time operating system has well-defined and fixed time constraints [15]. The system's operation is acceptable only if the task is accomplished within the defined constraints.

The procedure of sharing resources in OSes during which so much interference (allocation and deallocation of resources) occurs is another reason for OSes' unpredictability. Predictable latency is an important issue in resource management in cloud computing [3], [16]–[19].

The nature of multiprocessor architectures is associated with unpredictability. Utilizing a particular type of memory access is an example of resource sharing that increases unpredictability in an OS. There are two types of shared-memory multiprocessors: Uniform Memory Access (UMA) and Non-Uniform Memory Access (NUMA) [20] . In NUMA systems, each processor has special (efficient) access to a memory section, called a NUMA node. For each process, memory can be allocated locally (from the local node) to the corresponding CPU [21]. This allocation process is called NUMA placement. If the local node can not fulfil the memory request, the kernel (Linux) will seek the remote options [22] . NUMA placement is a source of uncertainty, and some works such as [23] have proposed ML models to predict its influence on applications' performance.

Other causes of unpredictability, as described in [20] , are pipeline optimization, cache interference, which increase memory traffic, and the NP-hard problem of scheduling in assigning tasks to processors. Also, as mentioned before, many OS algorithms consist of heuristics. For example, when all the page frames in memory are allocated, the kernel must decide which old pages can be freed and prevent them from being used by new processes. For this purpose, heuristics like the Least Recently Used (LRU) are being used [24]. LRU is originally a memory frame replacement policy (usually in cache) in which the least recently used items will be abandoned first. The implementation of this idea might seem straightforward, while keeping the track of least recently used items may become expensive. However, still designing an acceptable page frame reclaiming method is a trial and error job. Thus, it becomes a situation where the system becomes unpredictable; therefore, a perfect candidate for getting help from AI techniques such as "Reinforcement Learning (RL)." In RL, we examine how an agent can learn from success and failure, from reward and punishment [25] [15]. To follow a supervised learning approach, knowledge about the correct move of an agent in every step is mandatory. Such information is seldom available in complex environments like operating systems. Hence, a reward/punishment mechanism and a policy to maximize the expected reward would be feasible for such environments. The task of reinforcement learning is to use observed rewards to learn an optimal (or nearly optimal) policy for the environment [15]. In the next section, we investigate how ML can be beneficial to improve OS tasks.

## III. What is Data or the Problem Definition

To look at the operating system as a case study for applying artificial intelligence and machine learning, we must first define the problem accordingly. One way is to look at an operating system as a dynamic environment that contains a few agents, limitations and rules [15]. In this paper's definition, the OS kernel is the agent, its effectors are the responses that it gets to the system calls coming from the environment and the changes that it can cause in the policies and the limitations of the environment or on other agents. The kernel also precepts the environment via analysing the received system calls and interrupts. The kernel has algorithms to maintain the environment's stability, efficiency, security and fairness for other agents: hardware and processes. However, the vast difference among process agents sometimes jeopardizes the smooth activity of the kernel to manage the environment, requiring taking challenging (intelligent) decisions.

## IV. How OSes Benefit from ML

As mentioned above, several aspects of OSes are the candidates to apply AI, particularly ML, to enhance performance. Transforming OS's deterministic nature into a more flexible and dynamic form can maximize the exploitation of hardware and energy, for example, by reducing the idle times of the CPU. One of the features of ML is the ability to use previous experiences for future predictions. In this case, OSes with a massive amount of processes (threads and services) that continuously start and stop running can benefit from the processes' previous execution behaviour to manage them more optimally, for example, via allocating resources to them more intelligently according to their patterns of usage.

OSes, along with other systems like compilers, are full of heuristics that are hard-coded into their programs. Such heuristics are written concretely and cannot adapt to changes in usage [26]. This adaptation requires mechanisms to use the previous knowledge and patterns of users' behaviours to find the best configurations for timing operations, such as thread scheduling, memory paging, and the frequency of flushing buffer cache. Additionally, determining the optimal sizes for different containers such as buffer cache in storage caching is another operation that can be improved utilizing ML. In addition to learning the optimal configurations for timing and sizing activities, an OS can learn how to allocate memory spaces to applications or hard disk space to files in different situations, which is traditionally done according to predefined policies. Scheduling is one of the essential tasks of an OS, and it also has been one of the well-researched and implemented areas in Artificial Intelligence literature. The OS schedules the accessibility of the CPU to process threads, and its goal is to distribute the CPU time fairly and optimally among all threads running in the system. Other tasks, such as scheduling different network and storage queues, are of the same type. There are deterministic algorithms defined in OSes to manage the scheduling [1]. Applying AI methods to learn the CPU usage pattern by different processes and applications, we may achieve better and more adaptive scheduling mechanisms.

Also, Since working with OSes is a continuous activity and is in direct contact with humans to fulfil their tremendous requests, online approaches seem beneficial. For example, in space allocation tasks, exploiting Reinforcement Learning is a proper method of ML. This method can alleviate the extra time and energy consumption, and performance penalties that the system suffers in case of wrong decisions [1]. In the next sections, we provide some efforts reported in the literature of applying ML methods to enhance OS tasks.

### A. ML in I/O Scheduling and Latency Management

To make I/O devices (e.g., disks) more compatible with modern processors, I/O schedulers must be in their most optimal form to eliminate the processing requests' overhead. Therefore, I/O scheduling is an important task of OSes in which AI can be used. In [27], a self-learning I/O scheduling scheme is proposed to classify different types of requests or workloads in run-time. To make such scheduling decisions, in the data collection phase, they gathered request features, including classes and sizes of the requests, number of processes, and inter-request distances between current and previous requests. Also, workload features, like the number of reads/writes, average request size, and the average number of processes, are collected. They found that Support Vector Machines (SVM) classification produces the lightest overhead. The whole ML system was implemented in Linux kernel 2.6.13 by modifying kernel I/O schedulers. However, the details of implementation are not provided. They found online learning more adaptive than offline learning.

Using their own in-kernel ML library, KMLIB, researchers in [28] employed a regression model to early-reject I/O requests with a high chance of not meeting the deadline. Learning data collected by producing random read and write operations with four threads on a 1 GB data set by running an I/O tester called FIO [29]. They modified the "mq-deadline" I/O scheduler in Linux Kernel 4.19.51 and adopted their ML library. The thresholded regression model's accuracy indicating whether an I/O request misses the deadline or not was 74.62%, which reduced the overall latency by 8%.

In a continuation of previous works like [30]–[32] which were based on some heuristics, the LinnOS introduced in [17] was an effort to deal with the problem of high latency in flash storage and SSDs by possessing a model that can learn the behaviour of the storage device. It can inform client applications about the anticipated per-I/O speed that the storage can provide for their current requests. LinnOS contains a computationally lightweight neural network to reduce the overheads of the learning and inferring phases. Each incoming request is classified either as a slow-speed or fast-speed.

In their online approach, if the latency is fast-speed, it will be passed to the storage, otherwise, with a slow-speed latency, the application will be informed, and the request will be denied. In this case, the application tries another storage node. The NN's input features extracted from the current and recent I/O requests are the number of I/Os in the queue when the new request arrives, the latency of the four most recently accomplished requests, and the number of pending I/Os at the time of arrival of that 4 I/Os. The output is the inference of the speed of the I/Os (slow or fast). Because of using SSDs, data gathering was not an issue. They collected data during busy hours to achieve more general and richer training data. The NN had three layers with linear neurons.

In LinnOS, the training phase is done in the user space using TensorFlow [33] then the weights are sent to NN running in kernel space after transforming them to integer values to compensate for the lack of support for floating-points in the kernel space. The NN is implemented in the block layer part of the Linux Kernel and needs 68 KB of kernel memory.

### B. ML in Scheduling

Scheduling algorithms are the heart of any OS. Their purpose is the fairness of time and memory allocation among all processes while the system operates optimally, and the processes finish their jobs as soon as possible. For example, this fair distribution of time among processes is done by Linux's Completely Fair Scheduler (CFS) in the Linux kernel. However, CFS has limitations in multicore environments where it becomes so complicated [34]. Researchers, in [35], used an ML method to balance the Linux kernel load on a multiprocessor computer. They modified a kernel function and embedded their Multi-Layer Perceptron (MLP) model with three layers for load balancing decisions. The forward phase of the MLP was implemented in C and contained floating-point computations. However, the data collection phase was implemented in a two-way fashion between kernel and userspace using some tools explained in their paper. The training set contains 500,000 records resulting from calling load-balancer in different levels of workloads. The same amount of data also was collected in different CPU load averages. Fifteen input features, including the combination of Idle time of the target CPU, NUMA node numbers, and running time of the process per core, were collected. However, in addition to the extra load of the MLP, they did not see any noticeable difference between the original Linux scheduler and the ML-based one.

The increasing usage of Deep Learning methods raised concerns about the Quality of Services of DL-based applications. For example, the CFS can not effectively handle the resulting excessive memory traffic caused by requests for DL-based services [36]. New strategies based on ML should be considered for this situation, and it is interesting because, here, ML can be used to enhance ML services.

Many deep learning methods and platforms are using server-less computation, where microservices play the computation role. With a considerable inclination towards microservices on the cloud in AI-based applications, particularly Deep Learning methods, resource scheduling among many micro-services, therefore the Quality of Service on a server is becoming more challenging for traditional schedulers designed for OSes [3]. Here the challenge is to find the most optimal allocation of resources to different micro-services so that the QoS maximizes. This problem is a searching problem.

The resources to be allocated among services include CPU cores, main memories, cache, and bandwidth. The number of micro-services fluctuates, and also, these services are computationally heavy and sensitive to the reduction of resources during their run-time. The traditional forms of resource management and scheduling of OSes are not adapted to the heavy computation needs of deep learning applications.

The OSML scheduling mechanism, introduced in [3] attempts to reach the best QoS for microservices. It uses ML to find the optimal amount of resources (cores and cache) that any micro-service requires (OAA: Optimal Allocation Area) with which the OS understands that extra resource is not needed by the service. It also prevents micro-services from facing a sharp reduction in their QoS due to loss of resources pre-empted by CPU for serving other microservices (RCliffs: Resource Cliffs). The models are designed on TensorFlow, and the whole mechanism runs in the userspace. They used four three-layered Multi-Layer Perceptrons (called Model-A, Model-B, and their shadows) and a modified version of Deep Q-Network [37], [38] (called Model-C). The Reinforcement Learning approach in Model-C was used to correct the wrong decisions in what Model-A and B propose for resource management. OSML acts as an alerting mechanism to the kernel scheduler.

In [39], the authors reported an experiment in which they exploited ML methods to predict CPU burst times for a process. Several CPU-bound programs, including matrix multiplication, sorting programs, recursive Fibonacci number generating programs, and random number generator programs, were analysed. They added two system calls to the Linux kernel for providing interoperability between the kernel space scheduler and the Decision Tree inference mechanism in the userspace. One of the system calls was responsible for taking the Special Time Slice (STS) classified by the C4.5 decision tree and set the time_slice variable of the process descriptor in the modified scheduler at the kernel level. Another system call was also used to take the "time-rewarded" process back to the normal condition to be allocated the time slice according to the system default task_timeslice() procedure.

*C. ML in Cache Management*

There are several caches in a computer system, such as CPU cache, web cache, file system cache. Each cache reduces the waiting time of the faster device for receiving data from the slower device. Another challenging job of an OS is to handle the caches (in virtual memory or file systems) to know what data should be cached, how much time the data should remain in the cache always to maintain only data on immediate demand in the cache and remove others [1][40]. Caches always try to maintain more data to help processes run faster [10]. However, the size of caches and the cache hit rates are essential criteria in the system's efficiency. Therefore, keeping the cache's size as minimal as possible while reducing the amount of removing operations and maximizing the cache hit rates are trade-offs in caching procedures. Learning the patterns of data usages by different processes or users can help the OSes to pick the data better to be evicted from the cache.

How a caching procedure operates at the OS level is currently hard-coded in the kernel, based on predefined assumptions about data and users' behaviour. But, the workloads in caching are not constant and face many fluctuations. Since caching occurs in different computer system levels, it is regarded as a "caching problem" and is worked on in other researches as a general problem to be solved.

For example, in [40], the authors investigated the improvement in online cache management by applying reinforcement learning based on the reward and impact that a caching decision has brought to the system's performance. As the authors reported, along with a better adaptation to workloads, the approach can provide a higher hit rate while minimizing the memory space needed. The method was implemented using TensorFlow. Although its results may be convincing for networking, it is mainly designed for user space applications, and the challenges for implementing in kernel space must be considered.

Other works such as [41] also looked at this as a general problem and only provided ML-based approaches from this perspective. On the other hand, some researchers tried to solve this problem at the hardware level [42], [43]. For example, in [43], researchers used an LSTM-RNN-based cache replacement mechanism. Using the insights from that, they could create a lighter SVM-based architecture at the hardware level to solve cache replacement. Their method gained better performance than previous heuristic methods such as LRU. Accordingly, applying ML algorithms in OS level cache management still needs more efforts from the researchers.

*D. ML in Malware Detection*

The malicious behaviour of processes can be determined by monitoring how the kernel operates as a final result of their requests. Like any other process, the malware also utilizes specific system calls. There are several works in which ML is used to identify malware by monitoring the system calls that they use. For example, in [44], for Android systems, they used system calls of and executable file as features since they demonstrate how program communicate with the kernel. They also weighted the system calls mostly invoked by each application to boost the discrimination method during generating their data set. Several classification methods were utilized to investigate whether an executable file is a malware or not. Although the approach showed a noticeable performance in identifying malware apps, the vulnerability against malicious learning data remains a severe challenge in this area. We will explain this issue in the following sections.

Also, increasing the security of devices that are not connected to the Internet or have no definitive way to remain up-to-date through getting security patches is a serious issue, particularly in embedded OS. Using ML as a dynamic way of controlling the behaviour of processes and requests by distinguishing the system's expected behaviour from the anomaly may help them stay secure independently. In [45], a host-based, run-time anomaly detection mechanism for Linux OSes is proposed to increase the firmware's security in embedded systems. Their mechanism consists of three parts in both kernel and user spaces. Deep Learning models, in their Exein ML Engine (MLE), were trained to learn the normal behaviour of the systems' processes.

They categorized processes into several classes and put a tag identifying that class into the corresponding executable files during the firmware build. Then, for each type, an ML model was designed. MLE is a userspace procedure with Convolutional Neural Networks trained based on the processes' behaviour at the kernel level.

"Anomaly score" as the model's output is the value used to indicate how unusual a process is acting. Suppose a process is labelled as malicious during execution time. In that case, another module, called Exein Linux Security Module (LSM) that works in direct contact with the kernel (kernel level), will be informed. Every system calls of a monitored process during its run-time is hooked and sent to LSM via a kernel module: The Exein Linux Kernel Module (LKM). Then LSM will extract each call's data such as file descriptors, name, paths, inode attributes, memory information, and permission attributes. This information will be sent to the MLE to label the behaviour of the process again via LKM. LSM will be informed of the MLE decision and performs accordingly to trust or distrust the process. Therefore, the LKM acts as an interface between the MLE in userspace and the LKM in kernel space.

Such a malware detection procedure adds a significant computational load on the OS, as the authors admitted. However, this load is mostly coming from the process monitoring part and the information sent and received by the three modules, which was alleviated by shrinking the whole process data to limited "snapshots" or only monitoring processes that communicate with the network as the main entrance for attacks. The learning phase of ML is done offline (before building the embedded firmware); however, CNN's inference load added to the OS also should be considered.

## V. COMPLEXITIES IN USING ML IN OS

In this section, we encounter some challenges in using ML models in the OS environment. Possible or tested solutions in the literature also are analysed. Here, we tried to add different aspects to those stated in [1].

### A. Implementation Constraints (in Kernel Space)

The particular challenge in manipulating the kernel space mechanisms is limited access to libraries and programming languages in this space. One solution is to make a two-way path between the kernel and userspace to benefit from the user space's resources, not an optimal approach. For example, The main problem with the two system calls used in [39] was extra transitions from user mode to kernel mode, and this is even more considerable if the learning method is online (adaptive) and needs to be trained regularly based on the new events in the system. The first system call was created due to the lack of enough flexibility in the kernel space. It is vital to consider the overload of such system calls in the system's overall performance evaluation, benefiting AI. Another solution is to implement the required tools in the kernel space.

KMLib, the ML library introduced in [28], contains the maths functions implemented from scratch to be used in the kernel space, mainly to deal with the inaccessibility of floating-point maths functions at this level. Based on the common preference of tensor computations in major deep-learning libraries such as TensorFlow or PyTorch [46], KMLib also used such an approach but obviously in a more constrained and less resource consuming manner. KMLib can operate in two modes: kernel mode and kernel-user memory-mapped shared mode. They performed the floating-point calculations in a code block started by kernel_fpu_begin and ended by kernel_fpu_end macros in the Kernel mode. Nevertheless, the code in this block must be small because of the overhead that it adds to the computations.

In contrast to the work in [28], researchers in [35] took a fixed-point approach in which a fixed number of bits is for storing the integer part, and the remaining bits are to store the decimal part of the number. Computation with such numbers is similar to that of integers.

Other techniques, such as the quantization approaches, exploited in TinyML [47], can be considered advantageous for this situation. The post-training quantization and quantization-aware-training approaches have shown practical benefit for reducing TensorFlow models' size and latency for the Internet of Things edge devices. Quantization consists of transforming 32-bit floating-point to 8-bit integers, which reduces model size and latency with a low impact on accuracy. However, such ideas are more practical for inference, not for training. Training a (Deep) learning model is challenging in low-resource environments lacking floating-point computational abilities [48], [49].

### B. Data Shortage and Data Gathering

Another issue in ML-based modifications in OS's low-level activities is related to data. The question is how the data should be gathered to be general enough for multi-purpose OSes' jobs. In [39], since the goal was to predict the best STS for each program, 84 data instances were generated by setting different STSs for five programs (with CPU-bound processes) and finding the best STS the one that minimizes the TaT (Turn-around-Time) of the processes of each program. Each data's features are several static and dynamic attributes of a process, mostly the sizes of different tables such as the hash table size and the program's size. A Genetic and an Exhaustive search are used to find the best set of features and the corresponding STS with which the process has the most efficient execution behaviour. Also, in online approaches collecting data is done by watching the system's state, then train a model regularly, which adds up more load on the system as we saw in [45].

### C. Many Parameters

Each OS contains hundreds to thousands of processes running and many resources managed by the kernel. Every modification in one part of the kernel must take into account other parts. For example, as in [39], the only evaluation criteria was the reduction of the TaT of one program being tested. However, the consequences of such a change in other parts of the OS are not evaluated.

### D. Computational Needs of ML Inference and Learning

Every AI algorithm also takes several computing resources. In the case of learning algorithms, if we assume the training phase as a separate procedure from the inference phase (which is not in online learning methods), still the inference phase takes some time to produce the results. Generally, to achieve a more reliable and more data-oriented application, we need a considerable amount of data and, consequently, more complicated learning models like those introduced in the Deep Learning literature, which also takes more time to infer. However, other DL approaches can be tried, such as few-shot learning, [28] in which the training and inference time and the amount of data can be reduced.

## E. Security Threats

Penetrating the isolated and tightly controlled environment of the kernel by outside data motivates malicious users to feed data that can mislead the learning models and, consequently, jeopardize the OS's logical operation. There are recommended solutions, such as using separate ML models for different applications. However, concerns remain for side-channel attacks [1]. Side-channel attacks are intimately related to patterns in the data gathered from a set of physically observable phenomena caused by the computation tasks in present microelectronic devices [50]. It can roughly be considered that these attacks are about the computer system's physical implementation rather than software implementations and vulnerabilities. In [39] as authors admitted, a simple threat was that their method did not take the security holes that are embedded into the kernel by their method since there were not any precautions to prevent a malicious user from creating a process that requests the maximum amount of STS (special_time_slice) for itself that can have excessive access to CPU.

Works such as [28] via implementing the whole ML into the kernel can eliminate such threat from the outside world with the penalty of more limited capabilities for making more powerful models. Additionally, every combination of basic kernel tasks with AI methods must consider the overloads that extra security precautions can cause.

## F. Interpretability (Explainability)

The ML methods' black-box nature makes their results unexplainable for the end-user and the experts. Explainable AI and recently, with larger and more complex learning models, explainable DL, have become an important topic in the literature [51], [52]. Therefore, in OS designed to have high predictability (in its traditional meaning and despite the potential vulnerability) and every response is based on a definitive algorithm, probing ML methods with little or no interpretability requires a change of perspective.

## VI. CONCLUSION AND FUTURE

In this paper, we reviewed some recent works on integrating ML methods into operating systems. According to these works, we also mentioned some challenges and possible solutions in reaching the idea of "Intelligent OS." As we listed, research is done to utilize ML to enhance the main tasks of an OS, such as process scheduling, memory allocation, I/O and cache management, and malware detection. These methods were implemented in user or kernel spaces or both. In user space, designers have access to tools such as TensorFlow to train powerful models, while the overhead was tremendous.

On the other hand, in-kernel methods must cope with the computational limitations. Security was another issue in methods that use both spaces. Security was also a problem in using user' (processes') data to create a data-driven operating system because of malicious users' threat.

Now, the question is: is it better to transform the operating systems to become more acknowledging toward AI methods, for example, by providing richer programming libraries in the kernel? Or try to solve the above challenges. Is it worth changing OSes in this way? Or we should stick to their deterministic nature and let the under-world of the kernel space remain as explainable as possible? Learning methods produce results that can not be explained, and there are efforts to make them as interpretable as possible to become more reliable. Although predictability is desirable in operating system design, explainability and interpretability of OS actions are also important. Making operating systems more thoughtful also will end in more computation for tasks that are already done with trial and error but faster. Dealing with these trade-offs requires lots of research and implementations.